\documentclass[10pt,twocolumn,letterpaper]{article}

\usepackage{iccv}              

\usepackage{times}
\usepackage{url}
\usepackage{epsfig}
\usepackage{graphicx}
\usepackage{amsmath}
\usepackage{amssymb}
\usepackage{ragged2e}
\usepackage{makecell}
\usepackage[dvipsnames]{xcolor}
\graphicspath{{./figures/}}

\usepackage{pifont}
\usepackage{bm}
\usepackage{cuted}
\usepackage{overpic}
\usepackage{multirow}
\usepackage{pifont}
\usepackage{color}
\usepackage{colortbl}
\usepackage{multirow, hhline}
\usepackage{dashrule}
\usepackage{setspace}
\usepackage{color}
\usepackage{enumerate}
\usepackage{capt-of}
\usepackage{stackengine}
\usepackage{bbding}
\usepackage{mdframed}
\usepackage{tikz}
\usepackage{soul}

\usepackage[accsupp]{axessibility} 

\definecolor{iccvblue}{rgb}{0.21,0.49,0.74}
\usepackage[pagebackref,breaklinks,colorlinks,allcolors=iccvblue]{hyperref}

\def\paperID{3254} 
\def\confName{ICCV}
\def\confYear{2025}

\title{Blind Video Super-Resolution based on Implicit Kernels}

\author{Qiang Zhu$^{1}$, Yuxuan Jiang$^{2}$, Shuyuan Zhu$^{1}\thanks{Corresponding author}$,  Fan Zhang$^{2}$, David Bull$^{2}$, Bing Zeng$^{1}$\\
$^{1}$University of Electronic Science and Technology of China, $^{2}$University of Bristol\\
{\tt\small \{zhuqiang@std.,eezsy@,eezeng@\}uestc.edu.cn,\{yuxuan.jiang,fan.zhang,dave.bull\}@bristol.ac.uk}
}

\begin{document}
\maketitle

\begin{abstract}

Blind video super-resolution (BVSR) is a low-level vision task which aims to generate high-resolution videos from low-resolution counterparts in unknown degradation scenarios. Existing approaches typically predict blur kernels that are spatially invariant in each video frame or even the entire video. These methods do not consider potential spatio-temporal varying degradations in videos, resulting in suboptimal BVSR performance. In this context, we propose a novel \textbf{BVSR} model based on \textbf{I}mplicit \textbf{K}ernels, \textbf{BVSR-IK}, which constructs a multi-scale kernel dictionary parameterized by implicit neural representations. It also employs a newly designed recurrent Transformer to predict the coefficient weights for accurate filtering in both frame correction and feature alignment. Experimental results have demonstrated the effectiveness of the proposed BVSR-IK, when compared with four state-of-the-art BVSR models on three commonly used datasets, with BVSR-IK outperforming the second best approach, FMA-Net, by up to 0.59 dB in PSNR. Source code will be available at \url{https://github.com/QZ1-boy/BVSR-IK}.
\end{abstract}

\section{Introduction}

Aiming to generate a high-resolution (HR) video from its low-resolution (LR) version, video super-resolution (VSR) has been widely applied in many scenarios, including video streaming~\cite{kang2023super,lin2023luma,afonso2018video}, surveillance~\cite{deshmukh2019fractional,yi2019progressive}, and remote sensing~\cite{xiao2023local,luo2017video}. It is noted that, in some practical cases, LR videos are potentially contaminated by unknown degradations, which makes the task even more challenging. On the contrary, existing generic VSR methods ~\cite{zhou2024video,xu2024enhancing,chan2021basicvsr,chan2022basicvsr++,zhu2024dvsrnet,zhu2024deep} are typically based on known blur kernels~\cite{chan2021basicvsr,chan2022basicvsr++} (e.g., Gaussian blur kernel) without specifically modeling them during the restoration process. As a result, these VSR methods are not able to effectively capture the intrinsic characteristics of degraded videos~\cite{chu2020learning} in those challenging cases, leading to suboptimal VSR performance.

\begin{figure}[htbp]
\centering
\includegraphics[width=0.98\linewidth]{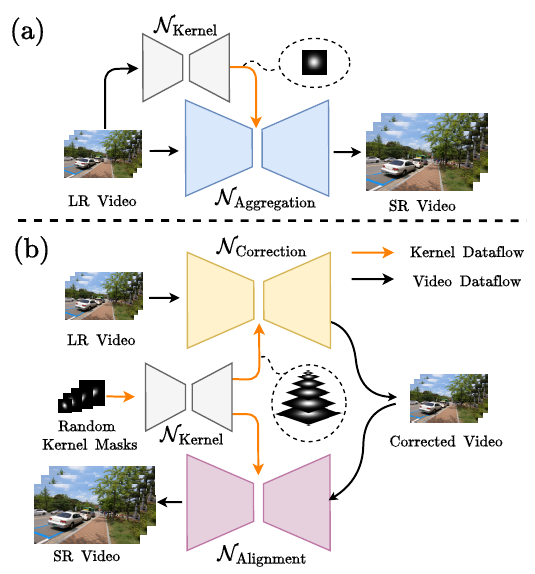}
\caption{Comparison between existing BVSR methods \cite{bai2024self,pan2021deep,xiao2023deep,youk2024fma}  and our BVSR-IK method. (a) Previous methods predict a single kernel and utilize it to perform super-resolution in an \textbf{Aggregation} manner. (b) BVSR-IK (ours) generates an INR-based multi-scale kernel dictionary and applies it in a \textbf{Correction}-\textbf{Alignment} manner for BVSR.}
\label{fig_Comp}
\vspace{-14pt}
\end{figure}

To address this issue, different blind VSR (BVSR) methods~\cite{pan2021deep,lee2021dynavsr,xiao2023deep,jeelani2023expanding,wang2023benchmark,yun2024kernel,bai2024self} have been proposed, which are designed to predict specific blur kernels or extract degradations in each inference case. Although these BVSR methods can offer promising super-resolution performance in unknown degradation scenarios, they are also associated with various limitations: (i) these approaches typically assume blur kernels are spatially invariant within each video frame or the whole video, while such an assumption is not applicable in many real cases where blur kernels are spatially variant due to motion and/or focus blurs; 
(ii) the temporal alignment modules within many BVSR methods~\cite{pan2021deep, xiao2023deep, yun2024kernel} assume that video frames share the same degradations~\cite{pan2021deep, xiao2023deep} - this ignores the temporal inconsistency of degradations among frames. In both cases, visual artifacts often exhibit (e.g. degradation and misalignment artifacts), in particular, when the degradations are complex.

In this context, this paper proposes \textbf{BVSR-IK}, a novel \textbf{BVSR} framework (as illustrated in \autoref{fig_Comp}) based on \textbf{I}mplicit \textbf{K}ernels, which generates a multi-scale kernel dictionary using implicit neural representations (INR) to efficiently parameterize multi-scale kernel atoms and implicitly model spatially varying degradations. Moreover, based on the implicit kernels, a new recurrent Transformer is designed to predict coefficient weights by capturing both short-term and long-term scale-aware information. These predicted coefficient weights are then employed for accurate filtering in two modules developed in this framework, Implicit Spatial Correction (ISC) and Implicit Temporal Alignment (ITA), which produce the final super-resolved video. The main contributions are summarized as follows:

\begin{itemize} 

\item We proposed \textbf{the first BVSR model based on implicit kernels}, which predicts multi-scale kernel representations for spatially varying degradations.

\item It is also the first time to employ a (modified) \textbf{recurrent Transformer for coefficient weight prediction}, which can capture sufficient spatial and temporal information.

\item \textbf{A new correction-alignment framework} is built for BVSR, based on the predicted implicit multi-scale kernels. Existing methods typically adopt an aggregation-based approach (as shown in \autoref{fig_Comp}).  

\end{itemize}

The proposed method has been extensively evaluated on three datasets, compared to four state-of-the-art (SOTA) BVSR methods. The results show that BVSR-IK has achieved consistent performance improvement over FMA-Net~\cite{youk2024fma} (the second best model in the experiment), with up to a 0.59 dB gain in PSNR. 

\section{Related Work} \label{RW}

\noindent {\bf{Blind Image Super-Resolution.}} Single image super-resolution~\cite{chen2023dual,chen2023activating,zhang2018residual,zhu2023attention,sun2023spatially,jiang2024mtkd} can offer satisfactory reconstruction performance when blur kernel is known, but perform poorly in unknown degradation scenarios. 
To address this problem, blind image super-resolution (BISR) methods have been proposed, and can be classified into two categories, kernel prediction (KP) based, and degradation extraction (DE) based. The former typically predicts blur kernels which are further used to guide the SR process. Some advanced techniques have been adopted for kernel estimation, including FKP \cite{liang2021flow} based on a normalizing flow-based kernel prior and MetaKernelGAN \cite{lee2024meta} through meta-learning enhanced kernel estimation adaptation. The DE-based methods ~\cite{liu2024degradation,wang2021unsupervised,xia2022knowledge,qiu2023dual} learn a degradation representation from the degraded image, with notable examples such as DASR~\cite{wang2021unsupervised}, CDFormer \cite{liu2024cdformer} and DiffTSR  \cite{zhang2024diffusion}.


\noindent {\bf{Blind Video Super-Resolution.}} Similar challenges are also associated with generic video super-resolution models~\cite{xue2019video,fuoli2019efficient,wang2019edvr,isobe2022look,chan2021basicvsr,chan2022basicvsr++}, which are based on known degradation conditions. Blind video super-resolution (BVSR) approaches have therefore been proposed ~\cite{pan2021deep,lee2021dynavsr,xiao2023deep,jeelani2023expanding,wang2023benchmark,yun2024kernel,bai2024self} based on different network structures and learning methodologies. DynaVSR~\cite{lee2021dynavsr} is one of the earliest works, which employs meta-learning to estimate a downscaled model and adapt to the current input, while DBVSR~\cite{pan2021deep} simultaneously estimates blur kernels, optical flows, and latent images for video restoration. More recently, BSVSR~\cite{xiao2023deep} has been proposed based on a multi-scale deformable convolution and deformable attention to enhance spatial details in a coarse-to-fine manner. Finally, FMA-Net~\cite{youk2024fma} is one of the latest contributions that employs 3D convolution layers for degradation learning, which has been reported to offer improved performance over other existing methods.


\noindent {\bf{Implicit Neural Representation (INR)}}
has emerged recently as a new compelling technique to effectively represent visual signals via coordinate-based multi-layer perceptrons (MLPs) \cite{saragadam2022miner}. 
It has been exploited in many application scenarios including 3D reconstruction \cite{chen2019learning, barron2021mip,lu2024unsigned}, image/video compression \cite{strumpler2022implicit, gao2025pnvc,kwan2024nvrc,kwan2023hinerv}, visual signal generation \cite{nam2022neural, xu2022signal}, and continuous image super-resolution \cite{chen2023cascaded, chen2021learning}. However, its application has not been investigated in the field of blind video super-resolution for kernel prediction.


\section{Proposed Method}

\begin{figure*}[!htbp]
\centering
\setlength{\abovecaptionskip}{0.2cm}
\includegraphics[width = 1\linewidth]{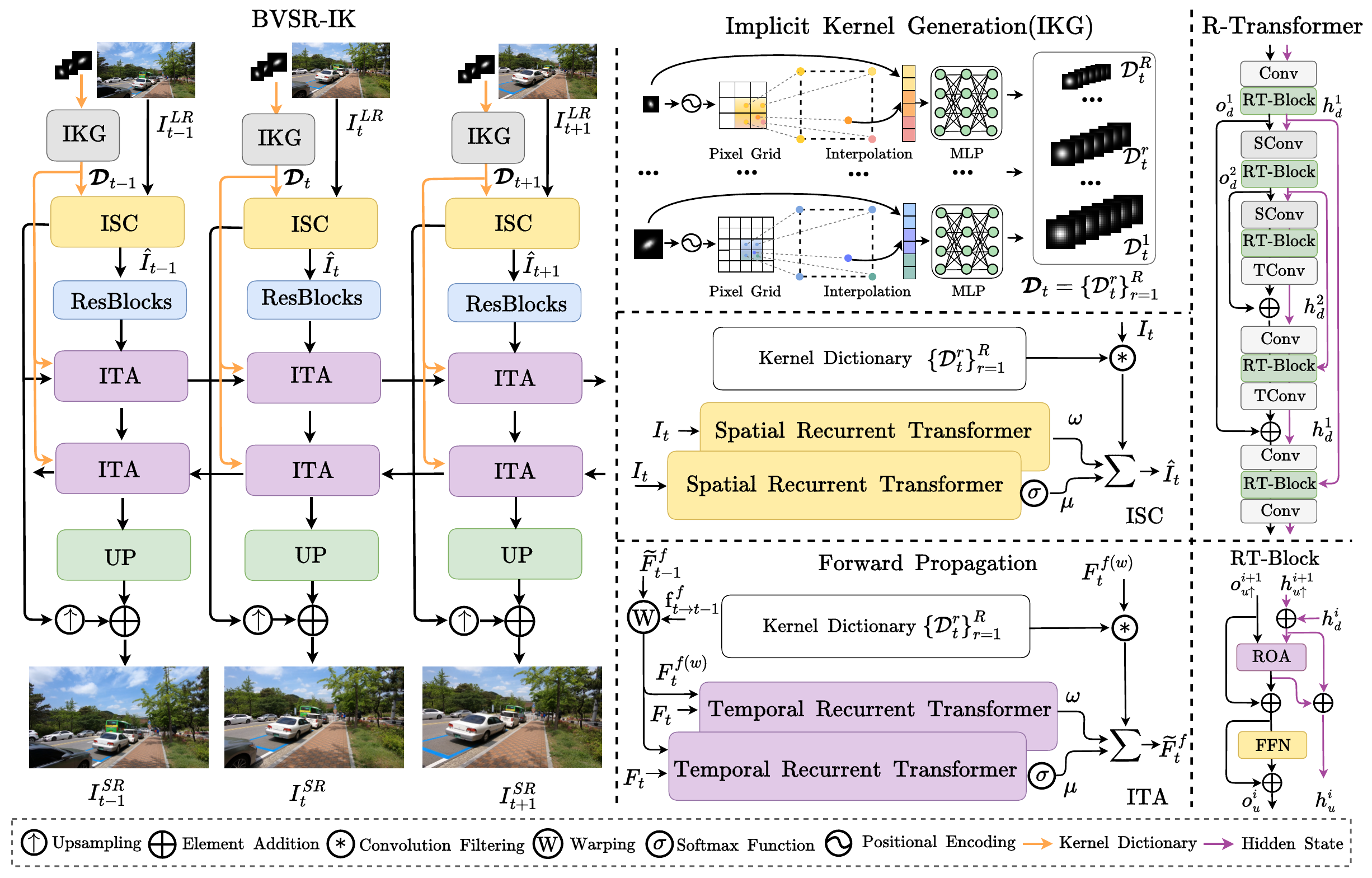}
\caption{The framework of the BVSR-IK model. The BVSR-IK model consists of three modules, i.e., ISC, ITA, and UP. Each LR video frame is first fed into the ISC module to generate its kernel dictionary and corrected frame. Then, the feature of the corrected frame extracted by residual blocks and the constructed kernel dictionary are fed into the ITA module to achieve the temporal feature alignment through bidirectional propagation. Finally, the aligned feature is fed into the UP module to generate the SR video frame.}
\vspace{-4pt}
\label{fig_pipline}
\end{figure*}

Let ${\bm{I}^{LR}}$ = $\{\emph{I}^{LR}_{t-M:t+M}\} \in \mathbb{R}^{(2M+1) \times H \times W \times 3}$ be a group of degraded LR video frames, the goal of BVSR is to generate its clear SR version $\bm{{I}}^{SR}$ = $\{\emph{I}^{SR}_{t-M:t+M}\}$ $\in\mathbb{R}^{(2M+1) \times sH \times sW \times 3}$, where $M$, $H$ and $W$ denote the temporal radius, height and width of the input frame group, and $s$ is the SR scale factor. As illustrated in~\autoref{fig_pipline}, the BVSR-IK framework consists of three primary components: Implicit Kernel Generation (IKG), Implicit Spatial Correlation (ISC) and Implicit Temporal Alignment (ITA) modules. Specifically, random kernel masks are first input into the IKG module to build an implicit multi-scale kernel dictionary $\boldsymbol{\mathcal{D}}_{t}$ for input video frame, $\emph{I}^{LR}_{t}$. They are then fed into the ISC module, which performs spatial correction for $\emph{I}^{LR}_{t}$. Through a few residual blocks (for feature extraction)  \cite{zhang2018image}, the resulting feature $\hat{F}_{t}$ of the corrected frame $\hat{I}_{t}$ together with those for two neighboring frames (indexed by $t-1$ and $t+1$) are sequentially taken by two ITA modules (also based on the multi-scale kernel dictionary $\boldsymbol{\mathcal{D}}_{t}$) to achieve inter-frame temporal alignment through bi-directional propagation, obtaining the forward aligned feature $\widetilde{F}^{f}_{t}$ and backward aligned feature $\widetilde{F}^{b}_{t}$. Finally, an UP-sampler (UP) is used to process the obtained backward aligned feature and produce the SR video frame ${I}^{SR}_t$. 

\subsection{Implicit Kernel Generation}  \label{INRDic}

 During the restoration process, the input frame $I^{LR}_t$ is processed a spatially varying deconvolution filtering $\mathcal{F}_{d}$ and a set of blur kernels $\mathcal{K}=\left\{\boldsymbol{k}_{i, j}\right\}_{i, j}$:
\begin{equation} \label{Decovn}
\mathcal{F}_{d}\left(I^{LR}_t ; \mathcal{K}\right)[i, j]=\sum_{x, y} \boldsymbol{k}_{i, j}[x, y] I^{LR}_t[i-x, j-y].
\end{equation}
Following \cite{quan2023single,guo2024spatially}, 
the blur kernel $\boldsymbol{k} = \sum_{n=1}^{N} \omega_n\boldsymbol{d}_{n}$ within a low-dimensional space can be expressed as a linear combination of atoms from a dictionary ${\mathcal{D}}$ = $[\boldsymbol{{d}}_1, \dots, \boldsymbol{{d}}_N ]$, where $\omega_n \in \mathbb{R}$ are the coefficient weights. 

In this work, to better handle spatially varying degradations and accurately approximate them using the minimum number of atoms, a multi-scale structure is usually introduced to construct the dictionary atoms. Specifically, we define a multi-scale kernel dictionary with $R$ sub-dictionaries, $\left\{{\mathcal{D}}^r\right\}_{r=1}^R$, each of which is defined as
${\mathcal{D}}^r=\left[\boldsymbol{d}_1^r, \cdots, \boldsymbol{d}_N^r\right] \subset \mathbb{R}^{M_r \times M_r}$. Here
 $\boldsymbol{d}_n^r$ denotes the atom at the $r$-th scale and $M_r$ is the size parameter of the atom, where $M_r$ gets smaller as $r$ increases. The per-pixel blur kernels are then expressed as:
\begin{equation}  \label{LineRep}
\boldsymbol{k}_{i, j} = \sum_{n=1}^{N} \omega_{n, i, j} \boldsymbol{d}_n^{r_{i, j}}\in \mathbb{R}^{M_{r_{i, j}} \times M_{r_{i, j}}},
\end{equation}
where $r_{i, j}$ denotes the scale index related to $\boldsymbol{k}_{i, j}$. Note that the atoms within the same sub-dictionary have a uniform size, while their sizes vary across different sub-dictionaries. 

Based on the linear representation in \autoref{LineRep} and the fact that $\boldsymbol{d}_n^{r_{i, j, j}} *$ $I^{LR}_t=\sum_r \delta\left(r-r_{i, j}\right) \boldsymbol{d}_n^r * I^{LR}_t$ with a Dirac delta function $\delta$,  \autoref{Decovn} is rewritten as:
\begin{align}
    \mathcal{F}_{d}(I^{LR}_t)  =\sum_{i, j} \mathbf{1}_{i, j} \odot\left(\sum_{n=1}^N \omega_{n, i, j} \boldsymbol{d}_n^{r_{i, j}} * I^{LR}_t\right) \nonumber \\
 =\sum_{i, j} \mathbf{1}_{i, j} \odot\left(\sum_{n=1}^N \sum_{r=1}^R \mu_{r, i, j} \omega_{n, i, j} \boldsymbol{d}_n^r * I^{LR}_t\right),
\label{Deconv1}
\end{align}
in which $\mathbf{1}_{i, j}$ represents an identity matrix, and $\mu_{r, i, j}=\delta\left(r-r_{i, j}\right)$. We can then perform spatially-varying deconvolution filtering by convolving each atom $\boldsymbol{d}_n^r$ in the dictionary with the input $I^{LR}_t$, and adding the resulting outputs to the spatially-varying weights $\mu_{r, i, j} \omega_{n, i, j}$. The weights $\mu_{r, i, j}$ and $\omega_{n, i, j}$ control the size (adaptive to scene depth) and shape (adaptive to image) of the corresponding blur kernel \cite{quan2023single}. The dictionary $\left\{{\mathcal{D}}^r\right\}_{r=1}^R$ used in \autoref{Deconv1} is generated by base atoms $\left\{\boldsymbol{d}_n\right\}_{n=1}^N$ with the smallest size.

Moreover, to better recover the base atoms, we use implicit neural representation (INR) to re-parameterize the multi-scale kernel dictionary. INR provides a more expressive way to generate multi-scale atoms that cover more frequencies than simple upsampling. Specifically, 
\begin{equation}
\boldsymbol{d}_n^r[x, y]=\phi_n(x, y),  [x, y] \in\left[1\cdots M_r\right] \times\left[1\cdots M_r\right],
\end{equation}
where $[x, y]$ is a spatial coordinate, and $\phi$ is an INR model implemented with a compact MLP. The INR model maps spatial coordinates to a continuous function that implicitly represents the kernel atom. The interpolation process is illustrated in \autoref{fig_pipline}. This function can be used to interpolate a kernel atom and create a dictionary of blur kernels with arbitrary scales. Based on this INR-based kernel representation, the learned kernel multi-scale dictionary of the input LR video frame $I^{LR}_{t}$ can be denoted as $\boldsymbol{\mathcal{D}_{t}}$ 
 = $\left\{{\mathcal{D}}_{t}^r\right\}_{r=1}^R$.

\subsection{Recurrent Transformer} 

To explore the scale-aware contextual information from input spatial frames or capture temporal features for generating the accuracy coefficient weights, we design a recurrent Transformer architecture consisting of recurrent Transformer blocks (RT-Block) in a multi-scale structure. This recurrent Transformer has been employed as the backbone to build both the ISC and ITA modules. 

As shown in \autoref{fig_pipline}, the recurrent Transformer involves operations at three different scales ($d_s=\{1,2,3\}$). At each scale level, two inputs, feature branch (black line) and hidden state branch (purple line), are fed into the convolution and RT-Block structure to obtain scale-aware feature and corresponding scale-aware hidden state. The stride convolution (SConv) and transposed convolution (TConv) are applied to downsample and upsample features. In the downsampling stage, RT-Block optimizes the input feature and hidden state and obtains the corresponding scale-aware feature $o_d^{i}$, $i\in\{1,2,3\}$ and the short-term hidden state $h_d^{i}$, $i\in\{1,2,3\}$. In the upsampling stage, the previous hidden states $h_d^{i}$, $i\in\{1,2\}$, and the long-term hidden states are passed to the upsampled same resolution RT-Block to achieve the optimization of the current short-term feature $o_u^{i}$, $i\in\{1,2\}$ and the long-term hidden state. The previous features $o_d^{i}$, $i\in\{1,2\}$ are added to $o_u^{i}$, $i\in\{1,2\}$ to enhance the scale-aware feature information. Based on this process, 
the coefficient weights can be effectively predicted.  

In each RT-Block, input feature and hidden state are optimized together based on the new recurrent optimization attention (ROA) and the feed-forward network (FFN) \cite{youk2024fma}. In the downsampling stage, no long-term hidden state is introduced into the ROA. As shown in \autoref{fig_pipline}, the previously upsampled feature $o^{i+1}_{u\uparrow} \in \mathbb{R}^{{H}\times{W}\times{C}}$ and the previously upsampled short-term hidden state $h^{i+1}_{u\uparrow} \in \mathbb{R}^{{H}\times{W}\times{C}}$ are fed into the ROA to interact them for scale-aware information optimization, where $H$,$W$,$C$ are the height, width and channel number of the feature. After that, the queries, keys and values are obtained by:
\begin{align}
Q_o, K_o, V_o=\mathrm{Split}(\mathcal{C}^{3}(o^{i-1}_{u\uparrow})),\\
Q_h, K_h, V_h=\mathrm{Split}(\mathcal{C}^{3}(h^{i-1}_{u\uparrow})),
\end{align}
in which $\mathcal{C}^{3}{(\cdot)}$ is a $3 \times 3$ convolution and $\mathrm{Split}$ is the channel split operation. We then reshape the query and key projections such that their dot-product interaction generates the output feature ${A} \in \mathbb{R}^{{H}\times{W}\times{C}}$. This is defined as:
\begin{equation}
A=\mathcal{C}^{1}( (V_o + V_h) \cdot \sigma( (K_o + K_h) \cdot (Q_o + Q_h) / \alpha)),
\end{equation}
where $\mathcal{C}^{1}{(\cdot)}$ is the $1\times1$ depth-wise convolution, $\sigma$ is the Softmax function and ${Q_o,O_h} \in \mathbb{R}^{{H} {W} \times {C}}, {K_o,K_h} \in \mathbb{R}^{{C} \times {H} {W}}$. ${V_o,V_h} \in \mathbb{R}^{{H} {W} \times {C}}$ are obtained after reshaping tensors from the original size $\mathbb{R}^{{H} \times {W} \times {C}}$. $\alpha$ is a learnable scaling parameter \cite{zamir2022restormer}. 

The last convolution of the recurrent Transformer is applied to convert the output feature to the weights $\boldsymbol{\mu}$ and $\boldsymbol{\omega}$. For predicting $\boldsymbol{\mu}$, the Softmax function is used to impose non-negativity and 1-normalization constraints. Such a soft relaxation of weights $\mu_{r, i,j}$ from 0, 1 to [0, 1] can improve the representation accuracy of blur kernels by including the use of additional atoms.

The recurrent Transformer is applied in the ISC module and the ITA module, denoting it as spatial recurrent Transformer (SRT) and temporal recurrent Transformer (TRT), respectively, to explore the spatial scale-aware information and temporal scale-aware information. 

\subsection{Implicit Spatial Correction}

The structure of the ISC module is illustrated in \autoref{fig_pipline}. To predict coefficient weights, our designed spatial recurrent Transformers are adopted to predict the coefficient weights $\boldsymbol{\omega}={SRT}(I_t)$ and $\boldsymbol{\mu}=\sigma({SRT}(I_t))$ for correcting input frames.  The input frame is fed into the feature branch and hidden state branch to apply spatial recurrent Transformers.

\subsection{Implicit Temporal Alignment}

In our model, the previously aligned features, optical flows, and constructed multi-scale kernel dictionary are fed into the ITA module through bidirectional propagation for temporal feature alignment. We present the process of the ITA module in forward propagation, which is illustrated in \autoref{fig_pipline}. The backward propagation has the same process.  Specifically, the forward aligned feature $\widetilde{F}^{f}_{t-1}$ is warped by an optical flow $\mathrm{f}^{f}_{t\rightarrow t-1}$ to obtain the forward warped feature  $F^{f(w)}_{t}$ = $\mathcal{W}(\widetilde{F}^{f}_{t-1}, \mathrm{f}^{f}_{t\rightarrow t-1})$, where $\mathcal{W}$ is the warping operation. The optical flow is generated by Spynet \cite{ranjan2017optical} from corrected frames $\mathrm{f}^{f}_{t\rightarrow t-1}$=$\mathrm{Spynet}(\hat{I}_{t},\hat{I}_{t-1})$. We then adopt our designed temporal recurrent Transformers to predict the coefficient weights $\boldsymbol{\omega} = \mathrm{TRT}(F^{f(w)}_{t}, F_t)$ and $\boldsymbol{\mu} = \sigma(\mathrm{TRT}(F^{f(w)}_{t}, F_t))$ for aligning temporal features.

\subsection{Loss Functions} \label{Loss}

To train the proposed BVSR-IK framework, we combine losses observing the ISC module and the overall performance. The total loss $\mathcal{L}$ is:
\begin{equation} \label{L_total}
\begin{aligned} 
\mathcal{L} & = \mathcal{L}_{R} + \lambda \mathcal{L}_{C} \\
& = \mathcal{L}_{Char}(\bm{I}^{SR},\bm{I}^{GT}) + \lambda \mathcal{L}_{Char}(\hat{\bm{I}}, \bm{I}^{DN}),
\end{aligned}
\end{equation}
where $\bm{I}^{GT}=\{\emph{I}^{GT}_{t-M:t+M}\}$ is the ground-truth HR video, and $\bm{I}^{DN}=\{\emph{I}^{DN}_{t-M:t+M}\}$ 
is the downsampled LR video from $\bm{I}^{GT}$. $\mathcal{L}_{Char}$ represents the Charbonnier loss \cite{lai2018fast}. $\lambda$ is the hyper-parameter. 


\section{Experimental Setup} 

\noindent \textbf{Implementation details.}\quad In the experiment, we set temporal radius $M$ = 2, scale factor $s$ = 4, the number of residual blocks $N_1$ = 8 for feature extraction. In the ISC module, we employ $R$ = 7 scales and $N$ = 8 INR-based atoms per scale. We use atoms with sizes 1$\times$1, 3$\times$3, \dots, 13$\times$13, where the 1$\times$1 atom is fixed as a delta kernel for modeling in-focus regions. The frequency parameters of the sine activation functions in all INR models are initialized by random sampling from [2, 16]. The UP module consists of $N_2$ = 13 residual blocks and two pixelshuffle layers for video restoration. 
During training, Adam optimizer~\cite{kingma2014adam} and the Cosine Annealing scheme~\cite{loshchilov2016sgdr}  are employed to optimize the BVSR-IK model. The initial learning rate is set to $1 \times 10^{-4}$. We trained our BVSR-IK model for 400 epochs. The training LR patch size is 256 $\times$ 256 and the mini-batch size is 7. We set $\lambda$ to 0.2 in our loss function. The models were implemented using PyTorch and trained with two NVIDIA GeForce RTX 3090 GPUs. 


\noindent {\bf{Datasets.}}
Following the previous works~\cite{bai2024self,liu2022learning,wang2019edvr}, we use REDS~\cite{nah2019ntire} as the training dataset to train our BVSR-IK model. To evaluate the performance of BVSR methods, three commonly used VSR testing datasets, including REDS4~\cite{nah2019ntire}, Vid4~\cite{liu2013bayesian} and UDM10 \cite{yi2019progressive}, are employed. 
We implemented two degradation scenarios, i.e., Gaussian blur and realistic motion blur, based on isotropic Gaussian blur kernel and realistic motion blur kernel, respectively. Previous generation of degraded videos \cite{pan2021deep, xiao2023deep, bai2024self} only used one kernel for one video, which hardly simulates the degradation change in videos. Therefore, we modified the generation method -  one kernel degrades one video frame for more accurate simulation. 
During training, each HR video frame is blurred and downsampled using a randomly generated kernel for both scenarios. Specifically, for Gaussian blur, the degradation is implemented by applying the randomly generated isotropic Gaussian kernel, where the range of the standard deviation $\sigma$ (of kernel) is configured as $\sigma\in$ [0.4, 2.0]. Moreover, we set the realistic motion blur scenario to simulate the realistic degradation scenario. The degradation is achieved by randomly generating realistic motion blur kernels. The Gaussian blur and realistic motion blur kernels are set to 13 $\times$ 13 kernel in terms of size.  We use the same approach to generate LR videos in the inference phase.


\begin{table*}[!t]
\centering
\renewcommand\arraystretch{1.0}
\footnotesize
\centering
\setlength{\tabcolsep}{3.7mm}
{\begin{tabular}{c|c|r|c|c|c}
\toprule
\multirow{2}[1]{*}{Scenarios}& \multirow{2}[1]{*}{Types}& \multirow{2}[1]{*}{Methods} & REDS4~\cite{nah2019ntire} & Vid4~\cite{liu2013bayesian}  & UDM10~\cite{yi2019progressive}  \\ 
 &  &   & {PSNR}$\uparrow$ / {SSIM}$\uparrow$ / tOF$\downarrow$  & {PSNR}$\uparrow$ / {SSIM}$\uparrow$ / tOF$\downarrow$   & {PSNR}$\uparrow$ / {SSIM}$\uparrow$ / tOF$\downarrow$  \\ 
\midrule
\multirow{10}[3]{*}{\makecell[c]{Gaussian \\ Blur}}
& \multicolumn{1}{c|}{\multirow{5}[1]{*}{\makecell[c]{Blind \\ SISR}}} 
&   DASR~\cite{wang2021unsupervised}  &  25.73 / 0.7062 / 4.17  &  22.98 / 0.6653 / 1.20  & 27.64 / 0.7661 / 1.30   \\ 
&   &   DCLS~\cite{luo2022deep}  & 26.19 / 0.7183 / 3.48   &  23.28 / 0.6754 / 1.12  & 28.06 / 0.7804 / 1.08 \\ 
&  &   DARSR~\cite{zhou2023learning}  & 25.26 / 0.6969 / 3.67   &  22.25 / 0.6413 / 1.73 &  27.81 / 0.7689 / 1.17 \\ 
&  & DIP-DKP~\cite{yang2024dynamic}  &  26.94 / 0.7286 / 3.25  &  23.50 / 0.6887 / 0.92  &  28.22 / 0.7898 / 0.96 \\ 
\cmidrule{2-6}
& \multicolumn{1}{c|}{\multirow{5}[1]{*}{\makecell*[c]{Blind \\ VSR}}} 
&   DBVSR~\cite{pan2021deep}    &   28.50 / 0.8071 / 2.03   &   23.98 / 0.6747 / 0.72  &  29.54 / 0.8227 / 0.96 \\
&    &{BSVSR}~\cite{xiao2023deep}   & 28.89 / 0.8113 / 1.91  & 24.15 / 0.6817 / 0.69  &  29.92 / 0.8423 / 0.89 \\ 
&  &   {Self-BVSR}~\cite{bai2024self}  &   29.08 / 0.8145 / 2.12 &  24.28 / 0.6918 / 0.65 &  30.08 / 0.8661 / 0.85  \\
&  & {FMA-Net}~\cite{youk2024fma}  & 29.11 / 0.8167 / 1.87  & 24.33 / 0.6925 / 0.68       & 30.19 / 0.8687 / 0.81 \\ 
&  &   \textbf{BVSR-IK (Ours)}  & \bf{29.50 / 0.8412 / 1.77}  &   \bf{24.48 / 0.7027 / 0.48}  &  \bf{30.21 / 0.8692 / 0.77} \\ 
\midrule
\multirow{10}[2]{*}{\makecell[c]{Realistic \\ Motion \\ Blur}}
& \multicolumn{1}{c|}{\multirow{5}[1]{*}{\makecell[c]{Blind \\ SISR}}} 
&   DASR~\cite{wang2021unsupervised}  &  25.49 / 0.7006 / 4.79   &  22.76 / 0.6133 / 1.43  &  27.43 / 0.7532 / 1.34 \\ 
&  &  DCLS~\cite{luo2022deep}  & 25.93 / 0.7082 / 3.73   &  22.98 / 0.6247 / 1.29  &  27.95 / 0.7769 / 1.06 \\ 
&  &  DARSR~\cite{zhou2023learning}  &  25.10 / 0.6958 / 4.27  &  22.06 / 0.6043 / 1.66  &  27.76 / 0.7668 / 1.19 \\
&  & DIP-DKP~\cite{yang2024dynamic}  & 26.75 / 0.7226 / 3.43  &  23.24 / 0.6488 / 1.09  &  28.16 / 0.7883 / 0.99 \\ 
\cmidrule{2-6}
& \multicolumn{1}{c|}{\multirow{5}[1]{*}{\makecell*[c]{ Blind \\ VSR}}} 
&   DBVSR~\cite{pan2021deep}  &   28.03 / 0.7882 / 2.83 &   23.73 / 0.6494 / 0.76  &  29.47 / 0.8216 / 0.94  \\ 
&  & {BSVSR}~\cite{xiao2023deep}  &28.58 / 0.8041 / 1.92  & 23.97 / 0.6559 / 0.69  &  29.82 / 0.8364 / 0.91 \\ 
&  &   {Self-BVSR}~\cite{bai2024self}   &    28.34 / 0.7926 / 2.21   &   24.04 / 0.6735 / 0.66  &  30.03 / 0.8568 / 0.87 \\ 
&  & {FMA-Net}~\cite{youk2024fma}  & 28.89 / 0.8094 / 1.89   & 24.16 / 0.6718 / 0.67  &  30.12 / 0.8675 / 0.83  \\ 
&  &   \textbf{BVSR-IK (Ours)}   &   \bf{29.48 / 0.8303 / 1.82}  &   \bf{24.52 / 0.7029 / 0.48}  &  \bf{30.27 / 0.8700 / 0.74} \\ 
\bottomrule	
\end{tabular}}
\caption{Quantitative evaluations by PSNR (dB), SSIM, and tOF on REDS4~\cite{nah2019ntire}, Vid4~\cite{liu2013bayesian}, and UDM10~\cite{yi2019progressive} datasets for Gaussian blur and realistic motion blur scenarios. The results are tested on the Y-channel. The best results in each case have been marked in \bf{bold}.} 
\vspace{-4pt}
\label{table1}
\end{table*}

\begin{figure*}[htbp]
\centering
\begin{minipage}[b]{0.210\linewidth}
\centering
\centerline{\epsfig{figure=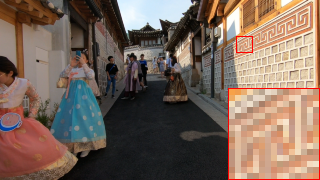,width=3.65cm}}
\footnotesize{REDS4\_011\_039 (Gaussian)}
\end{minipage}
\begin{minipage}[b]{0.118\linewidth}
\centering
\begin{overpic}[width=1\textwidth]{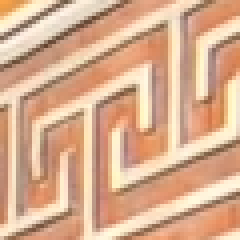}
\end{overpic}
\footnotesize{  GT  }
\end{minipage}
\begin{minipage}[b]{0.118\linewidth}
\centering
\begin{overpic}[width=1\textwidth]{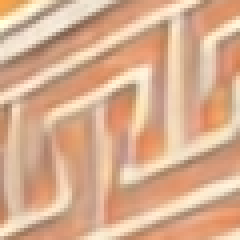} \end{overpic}
\footnotesize{  DBVSR~\cite{pan2021deep} }
\end{minipage}
\begin{minipage}[b]{0.118\linewidth}
\centering
\begin{overpic}[width=1\textwidth]{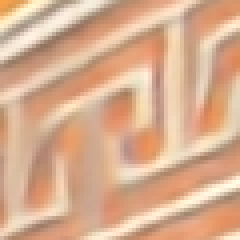} \end{overpic}
\footnotesize{  BSVSR~\cite{xiao2023deep}  }
\end{minipage}
\begin{minipage}[b]{0.118\linewidth}
\centering
\begin{overpic}[width=1\textwidth]{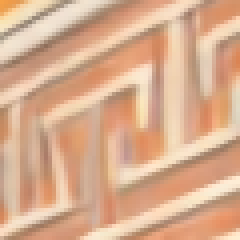}	
\end{overpic}
\footnotesize{  Self-BVSR~\cite{bai2024self} }
\end{minipage}
\begin{minipage}[b]{0.118\linewidth}
\centering
\begin{overpic}[width=1\textwidth]{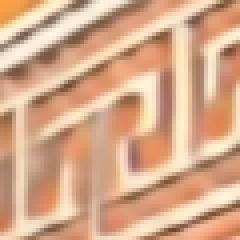}	
\end{overpic}
\footnotesize{ {FMA-Net}~\cite{youk2024fma} }
\end{minipage}
\begin{minipage}[b]{0.118\linewidth}
\centering
\begin{overpic}[width=1\textwidth]{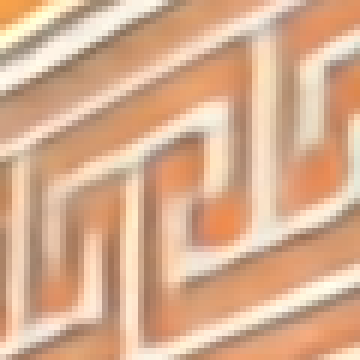}	
\end{overpic}
\footnotesize{ {BVSR-IK} (Ours) }
\end{minipage} \\

\begin{minipage}[b]{0.210\linewidth}
\centering
\centerline{\epsfig{figure=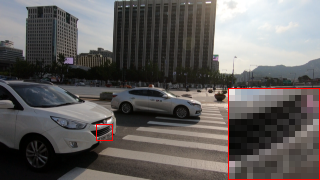,width=3.65cm}}
\footnotesize{ REDS4\_015\_049 (Realistic)}  
\end{minipage}
\begin{minipage}[b]{0.118\linewidth}
\centering
\begin{overpic}[width=1\textwidth]{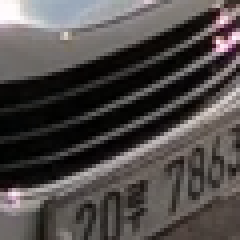}
\end{overpic}
\footnotesize{  GT  }
\end{minipage}
\begin{minipage}[b]{0.118\linewidth}
\centering
\begin{overpic}[width=1\textwidth]{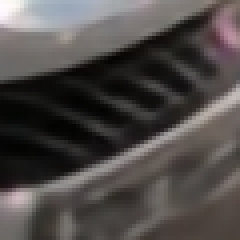}
\end{overpic}
\footnotesize{  DBVSR~\cite{pan2021deep} }
\end{minipage}
\begin{minipage}[b]{0.118\linewidth}
\centering
\begin{overpic}[width=1\textwidth]{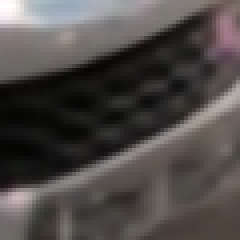}
\end{overpic}
\footnotesize{  BSVSR~\cite{xiao2023deep}  }
\end{minipage}
\begin{minipage}[b]{0.118\linewidth}
\centering
\begin{overpic}[width=1\textwidth]{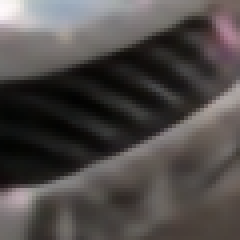}	
\end{overpic}
\footnotesize{  Self-BVSR~\cite{bai2024self} }
\end{minipage}
\begin{minipage}[b]{0.118\linewidth}
\centering
\begin{overpic}[width=1\textwidth]{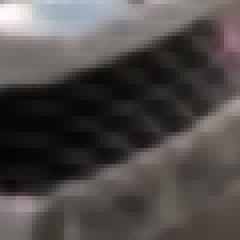}	
\end{overpic}
\footnotesize{ {FMA-Net}~\cite{youk2024fma} }
\end{minipage}
\begin{minipage}[b]{0.118\linewidth}
\centering
\begin{overpic}[width=1\textwidth]{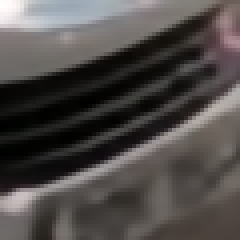}	
\end{overpic}
\footnotesize{ {BVSR-IK} (Ours)}
\end{minipage}

\begin{minipage}[b]{0.210\linewidth}
\centering
\centerline{\epsfig{figure=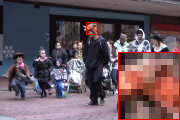,width=3.65cm}} 
\footnotesize{Vid4\_Walk\_019 (Gaussian)}  
\end{minipage}
\begin{minipage}[b]{0.118\linewidth}
\centering
\begin{overpic}[width=1\textwidth]{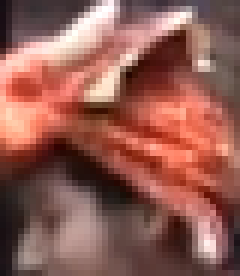}	
\end{overpic}
\footnotesize{  GT}
\end{minipage}
\begin{minipage}[b]{0.118\linewidth}
\centering
\begin{overpic}[width=1\textwidth]{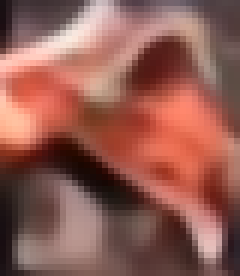}	
\end{overpic}
\footnotesize{  DBVSR~\cite{pan2021deep}}
\end{minipage}
\begin{minipage}[b]{0.118\linewidth}
\centering
\begin{overpic}[width=1\textwidth]{figures/DBVSR_gaussian_00000019.png}	
\end{overpic}
\footnotesize{  BSVSR~\cite{xiao2023deep}}
\end{minipage}
\begin{minipage}[b]{0.118\linewidth}
\centering
\begin{overpic}[width=1\textwidth]{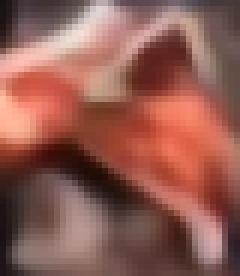}	
\end{overpic}
\footnotesize{  Self-BVSR~\cite{bai2024self} }
\end{minipage}
\begin{minipage}[b]{0.118\linewidth}
\centering
\begin{overpic}[width=1\textwidth]{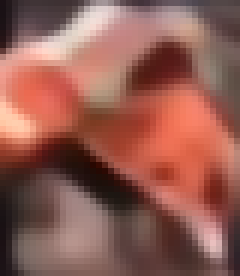}
\end{overpic}
\footnotesize{ {FMA-Net}~\cite{youk2024fma} }
\end{minipage}
\begin{minipage}[b]{0.118\linewidth}
\centering
\begin{overpic}[width=1\textwidth]{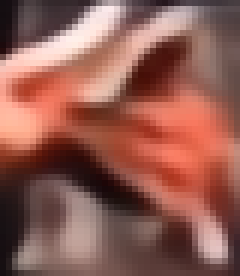}	
\end{overpic}
\footnotesize{ BVSR-IK (Ours) }
\end{minipage}  

\begin{minipage}[b]{0.210\linewidth}
\centering
\centerline{\epsfig{figure=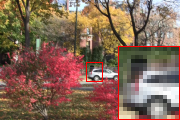,width=3.65cm}}  %
\footnotesize{ Vid4\_Foliage\_030 (Realistic)}  
\end{minipage}
\begin{minipage}[b]{0.118\linewidth}
\centering
\begin{overpic}[width=1\textwidth]{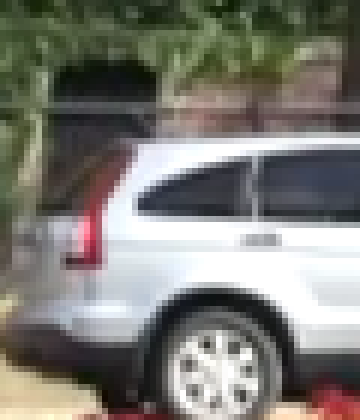}	
\end{overpic}
\footnotesize{  GT  }
\end{minipage}
\begin{minipage}[b]{0.118\linewidth}
\centering
\begin{overpic}[width=1\textwidth]{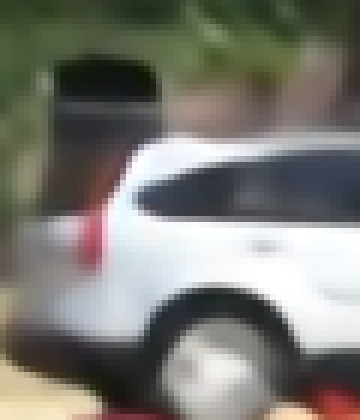}
\end{overpic}
\footnotesize{  DBVSR~\cite{pan2021deep} }
\end{minipage}
\begin{minipage}[b]{0.118\linewidth}
\centering
\begin{overpic}[width=1\textwidth]{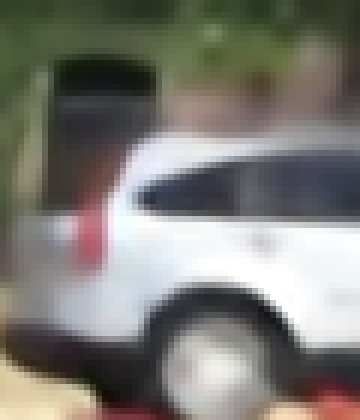}	
\end{overpic}
\footnotesize{  BSVSR~\cite{xiao2023deep}  }
\end{minipage}
\begin{minipage}[b]{0.118\linewidth}
\centering
\begin{overpic}[width=1\textwidth]{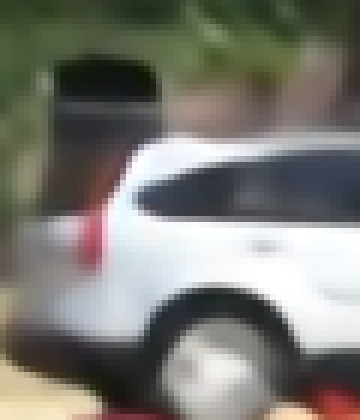}
\end{overpic}
\footnotesize{  Self-BVSR~\cite{bai2024self} }
\end{minipage}
\begin{minipage}[b]{0.118\linewidth}
\centering
\begin{overpic}[width=1\textwidth]{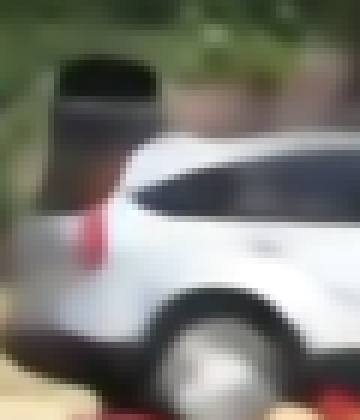}	
\end{overpic}
\footnotesize{ {FMA-Net}~\cite{youk2024fma} }
\end{minipage}
\begin{minipage}[b]{0.118\linewidth}
\centering
\begin{overpic}[width=1\textwidth]{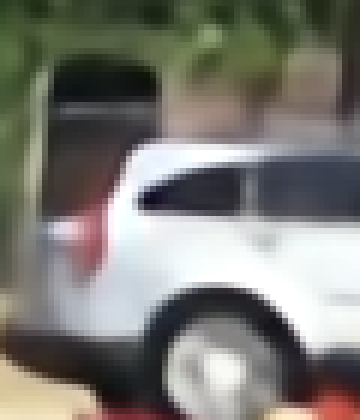}	 
\end{overpic}
\footnotesize{   {BVSR-IK} (Ours)  }
\end{minipage}

\begin{minipage}[b]{0.210\linewidth}
\centering
\centerline{\epsfig{figure=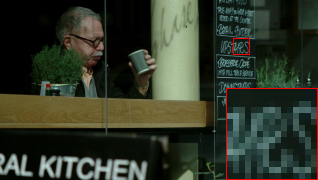,width=3.65cm}}  %
\footnotesize{ UDM10\_Caffe\_020 (Gaussian)}  
\end{minipage}
\begin{minipage}[b]{0.118\linewidth}
\centering
\begin{overpic}[width=1\textwidth]{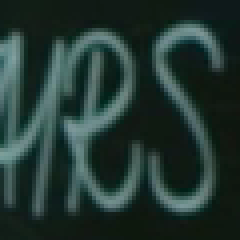}	
\end{overpic}
\footnotesize{  GT  }
\end{minipage}
\begin{minipage}[b]{0.118\linewidth}
\centering
\begin{overpic}[width=1\textwidth]{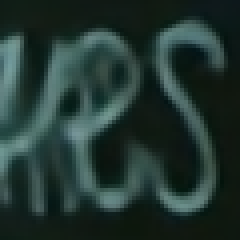}
\end{overpic}
\footnotesize{  DBVSR~\cite{pan2021deep} }
\end{minipage}
\begin{minipage}[b]{0.118\linewidth}
\centering
\begin{overpic}[width=1\textwidth]{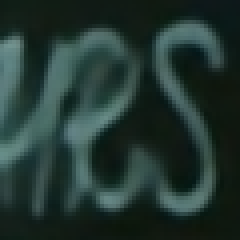}	
\end{overpic}
\footnotesize{  BSVSR~\cite{xiao2023deep}  }
\end{minipage}
\begin{minipage}[b]{0.118\linewidth}
\centering
\begin{overpic}[width=1\textwidth]{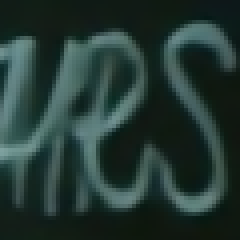}
\end{overpic}
\footnotesize{  Self-BVSR~\cite{bai2024self} }
\end{minipage}
\begin{minipage}[b]{0.118\linewidth}
\centering
\begin{overpic}[width=1\textwidth]{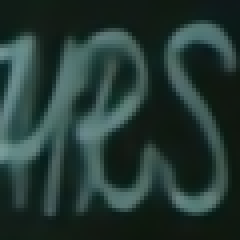}	
\end{overpic}
\footnotesize{ {FMA-Net}~\cite{youk2024fma} }
\end{minipage}
\begin{minipage}[b]{0.118\linewidth}
\centering
\begin{overpic}[width=1\textwidth]{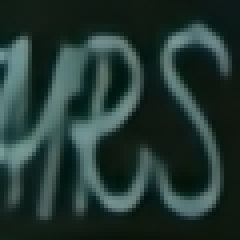}	 
\end{overpic}
\footnotesize{   {BVSR-IK} (Ours) }
\end{minipage}

\begin{minipage}[b]{0.210\linewidth}
\centering
\centerline{\epsfig{figure=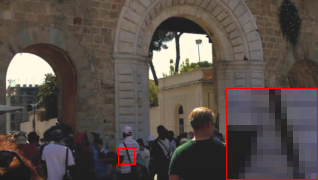,width=3.65cm}}  %
\footnotesize{UDM10\_Archpeople\_006  \\ (Realistic)}  
\end{minipage}
\begin{minipage}[b]{0.118\linewidth}
\centering
\begin{overpic}[width=1\textwidth]{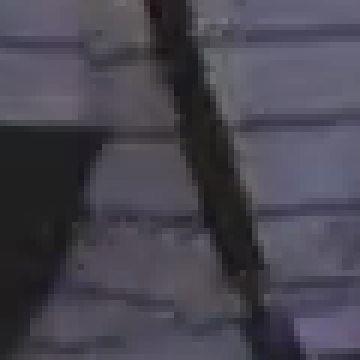}	
\end{overpic}
\footnotesize{  GT \\ \ }
\end{minipage}
\begin{minipage}[b]{0.118\linewidth}
\centering
\begin{overpic}[width=1\textwidth]{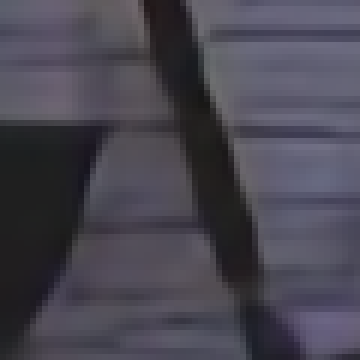}
\end{overpic}
\footnotesize{  DBVSR~\cite{pan2021deep}  \\ \ }
\end{minipage}
\begin{minipage}[b]{0.118\linewidth}
\centering
\begin{overpic}[width=1\textwidth]{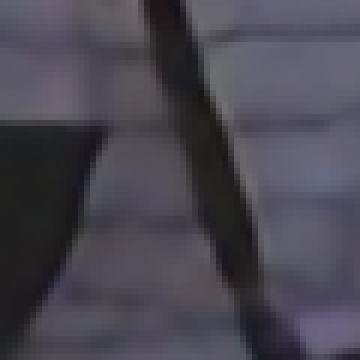}	
\end{overpic}
\footnotesize{  BSVSR~\cite{xiao2023deep}  \\ \ }
\end{minipage}
\begin{minipage}[b]{0.118\linewidth}
\centering
\begin{overpic}[width=1\textwidth]{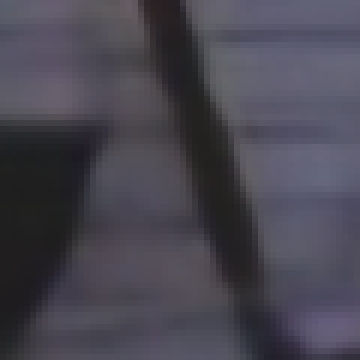}
\end{overpic}
\footnotesize{  Self-BVSR~\cite{bai2024self}  \\ \ }
\end{minipage}
\begin{minipage}[b]{0.118\linewidth}
\centering
\begin{overpic}[width=1\textwidth]{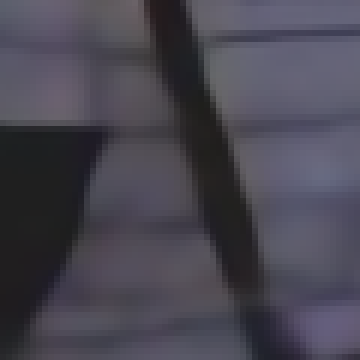}	
\end{overpic}
\footnotesize{ {FMA-Net}~\cite{youk2024fma}  \\ \ }
\end{minipage}
\begin{minipage}[b]{0.118\linewidth}
\centering
\begin{overpic}[width=1\textwidth]{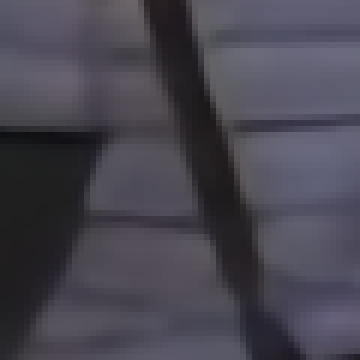}	 
\end{overpic}
\footnotesize{   {BVSR-IK} (Ours) \\ \  }
\end{minipage}
{\caption{{Visual results on REDS4~\cite{nah2019ntire}, Vid4~\cite{liu2013bayesian} and UDM10~\cite{yi2019progressive} datasets for Gaussian blur and realistic motion blur scenarios.} }
\vspace{4pt}
\label{fig_visual}}
\end{figure*}

\noindent {\bf{Evaluation metrics.}}
Aligned with the literature, PSNR, and SSIM~\cite{wang2004image} have been adopted to evaluate the restoration quality, and tOF~\cite{chu2020learning} is used to measure the temporal consistency of restored videos. Model size, FLOPs, and runtime are also calculated to estimate model complexity.


\noindent {\bf{Benchmark methods.}}
Eight SOTA BISR and BVSR methods have been used to benchmark the performance of the proposed BVSR-IK. These include four BISR models, DASR~\cite{wang2021unsupervised}, DIP-DKP~\cite{yang2024dynamic}, DCLS~\cite{luo2022deep} and DARSR~\cite{zhou2023learning}, and four BVSR approaches, DBVSR~\cite{pan2021deep}, BSVSR~\cite{xiao2023deep}, Self-BVSR~\cite{bai2024self}, FMA-Net~\cite{youk2024fma}. Due to the introduction of a new degradation generation method, we retrained these models based on their open-source implementations to obtain corresponding results.

\begin{table}[!t]
\footnotesize
\setlength{\tabcolsep}{1.20mm}
\renewcommand\arraystretch{1.0}
\centering
\begin{tabular}{r|c|c|c|ccc}
\toprule
Methods  & Frame  & Param.(M) $\downarrow$ &  FLOPs(G) $\downarrow$   & Runtime(s) $\downarrow$     \\
\midrule
DBVSR~\cite{pan2021deep}       &  5   & 14.10  & 158.69  & 0.86  \\
BSVSR~\cite{xiao2023deep}      &  5   &20.83   & 489.16  & 1.81  \\ 
Self-BVSR~\cite{bai2024self}   &  5   &18.20   & 281.45  & 1.29 \\ 
{FMA-Net}~\cite{youk2024fma}   &  5   & 9.94   & 564.42  & 2.07 \\
\midrule
\textbf{BVSR-IK (Ours)}   &  5   & 9.30  & 459.23  & 1.58 \\ 
\bottomrule
\end{tabular}
\caption{Comparisons of parameters and complexity. The FLOPs and Runtime are computed with  64$\times$64 LR video resolution.}
\label{table2}
\end{table}

\section{Results and Discussion}

\subsection{Overall Performance}

\noindent {\bf{Quantitative results.}} 
\autoref{table1} summarizes the results of the proposed and benchmark methods on three datasets for Gaussian blur and realistic motion blur scenarios. It can be observed that our BVSR-IK model achieves consistent and significant performance improvement over all the other BISR and BVSR models, with a 0.39 dB PSNR gain compared to the second best performer FMA-Net~\cite{youk2024fma}, for the Gaussian blur scenario, and 0.59 dB for the realistic motion blur scenario, both on the REDS4 testing dataset.


\noindent{\bf{Qualitative results.}} 
\autoref{fig_visual} provides visual comparison examples for four BVSR methods on three testing datasets under two degradation scenarios. It is evident that the reconstructed results produced by BVSR-IK exhibit better perceptual quality with fewer degradations and finer details compared to other benchmarks. 


\noindent{\bf{Model complexity.}} 
We evaluate the model complexity of four BVSR methods in terms of FLOPs, runtime, and model size on the REDS4 dataset for the realistic motion blur scenario. Specifically, we implement DBVSR~\cite{pan2021deep}, BSVSR~\cite{xiao2023deep}, Self-BVSR~\cite{bai2024self}, and FMA-Net~\cite{youk2024fma}, with $M$=2 temporal radius to achieve a fair comparison. The FLOPs and runtime are measured for 64$\times$64 LR video frames on a NVIDIA GeForce RTX 3090 GPU for generating HR videos. All the complexity figures are summarized in~\autoref{table2}. Our BVSR-IK model has the minimal model parameters and competitive complexity. Compared with the current SOTA method, FMA-Net~\cite{youk2024fma}, our method has reduced 0.64M parameters and 105.19G FLOPs complexity and obtained a significantly better performance gain.

\subsection{Ablation Study} \label{Abla}

To further verify the effectiveness of the main contributions in this work, we have designed an ablation study based on the REDS4 dataset for the realistic motion blur scenario. The corresponding results are provided in \autoref{Tab_Module}. 

\begin{table}[t]
\setlength{\tabcolsep}{0.50mm}
\renewcommand\arraystretch{1.08}
\centering
\resizebox{\linewidth}{!}{\begin{tabular}{l|c|c|c|cccc}
\toprule
Models  & { PSNR$\uparrow$ / SSIM$\uparrow$ / tOF$\downarrow$}  & {Param.(M)$\downarrow$}  &{FLOPs(G)$\downarrow$}  &{R.T.(s)$\downarrow$} \\
\midrule
\multicolumn{5}{c}{(a) Ablation Study of Modules and Loss} \\
\midrule
(v1.1) \emph{w/o} ISC      &   29.18 / 0.8128 / 1.87    &  8.12  & 447.45  &  1.42   \\
(v1.2) \emph{w/o} ITA      &  28.86 / 0.8089 / 1.92    &  5.50  & 373.51  &  0.97   \\
(v1.3) \emph{w/o} Rec      &   29.20 / 0.8139 / 1.86   & 8.55  & 435.13  &  1.48  \\
(v1.4) \emph{w/o} Bi-dir   &   29.29 / 0.8158 / 1.85   &  7.40  & 346.16  &  0.82   \\
(v1.5) \emph{w/o} $\mathcal{L}_{C}$  &  29.40 / 0.8289 / 1.84   &  9.30  &   459.23  &   1.58  \\
\midrule
\multicolumn{5}{c}{ (b) Comparison of Different Kernel Prediction (KP) with IK } \\
\midrule
(v2.1)  \cite{bai2024self}'s KP    & 29.16 / 0.8127 / 1.95  & 8.71  & 423.65 & 1.45 \\
(v2.2) \cite{youk2024fma}'s KP     & 29.35 / 0.8278 / 1.88  & 9.24  & 484.26 & 1.78 \\
(v2.3) DKP \cite{hussein2020correction}  & 28.92 / 0.8093 / 2.03  & 6.85 & 394.26 & 1.27  \\
\midrule
\multicolumn{5}{c}{ (c) Comparison of Different Correction Filters with ISC } \\
\midrule
(v3.1) Corr Filter \cite{hussein2020correction}  & 29.11 / 0.8107 / 1.92  & 8.21 & 451.21 & 1.45 \\
(v3.2) Corr Filter \cite{zhou2023learning}       & 29.20 / 0.8140 / 1.89  & 8.79  & 468.92 & 1.51 \\
(v3.3) MISCFilter \cite{liu2024motion}           & 29.35 / 0.8284 / 1.88  & 9.36 & 463.10 & 1.60 \\
\midrule
\multicolumn{5}{c}{ (d) Comparison of Different Alignment Modules with ITA } \\
\midrule
(v4.1) Flow \cite{chan2021basicvsr}      & 28.84 / 0.8082 / 1.98  & 5.57 & 373.60 & 0.99 \\
(v4.2) DCN  \cite{tian2020tdan}          & 29.08 / 0.8106 / 1.91  & 6.27  & 394.56 & 1.23 \\
(v4.3)  FGDA \cite{chan2022basicvsr++}   & 29.45 / 0.8309 / 1.86  & 9.58 & 467.32 & 1.66 \\
(v4.4)  FGDF \cite{youk2024fma}          & 29.21 / 0.8167 / 1.89  & 6.12 & 386.32 & 1.06 \\
\midrule
\textbf{BVSR-IK (Ours) }   &  29.48 / 0.8303 / 1.82   &  9.30  &   459.23  &   1.58   \\
\bottomrule
\end{tabular}}
\caption{{Results of the ablation study.}} \label{Tab_Module}
\end{table}

\begin{figure*}[htbp] 
\centering
\begin{minipage}[b]{0.19\textwidth}
\centering
\includegraphics[width=1.0\textwidth]{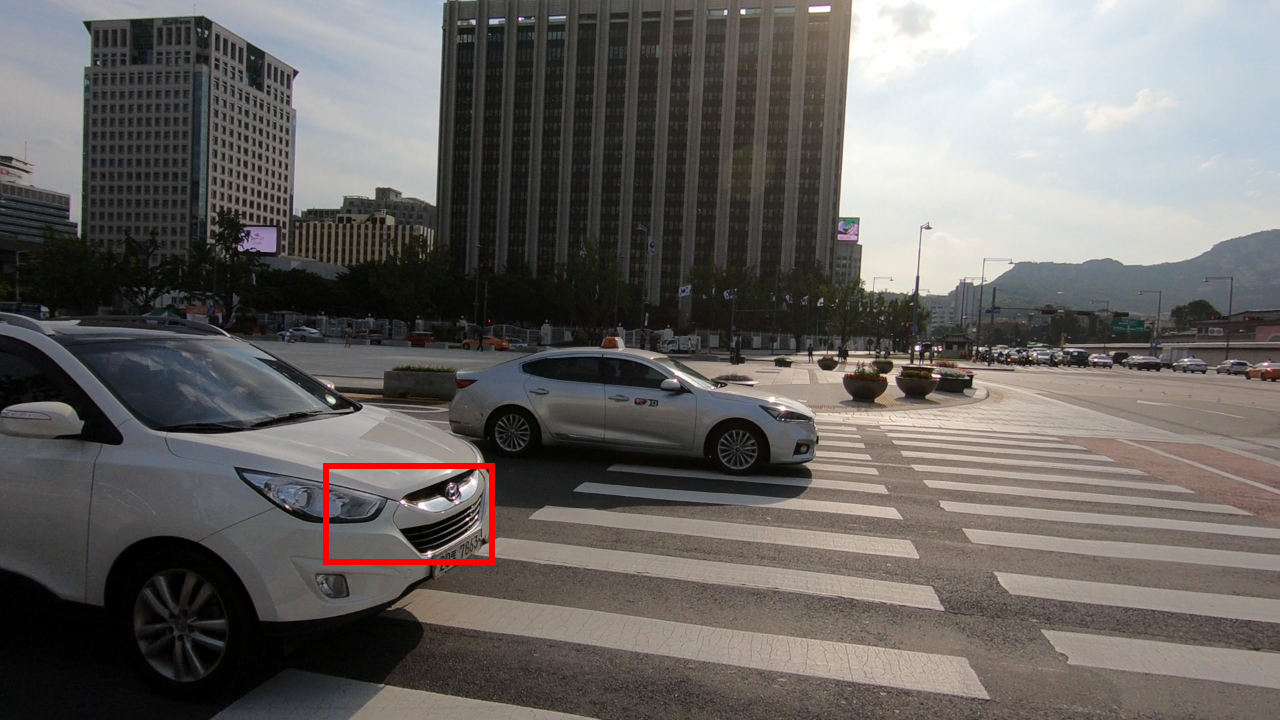}
\footnotesize{Reference Frame ($\emph{015\_050}$)}
\end{minipage}
\begin{minipage}[b]{0.19\textwidth}
\centering
\includegraphics[width=1.0\textwidth]{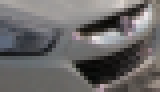}
\footnotesize{LR Frame }
\end{minipage}
\begin{minipage}[b]{0.19\textwidth}
\centering
\includegraphics[width=1.0\textwidth]{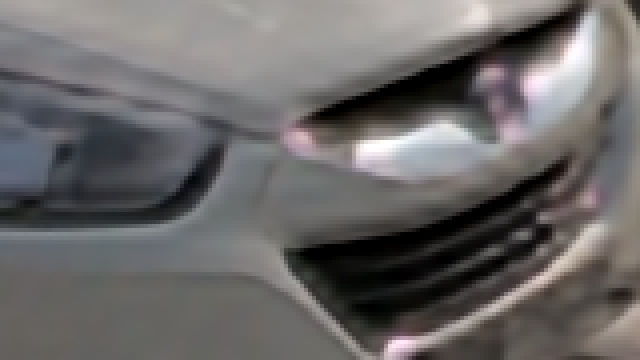}
\footnotesize{ (v1.1) w/o ISC}
\end{minipage}
\begin{minipage}[b]{0.19\textwidth}
\centering
\includegraphics[width=1.0\textwidth]{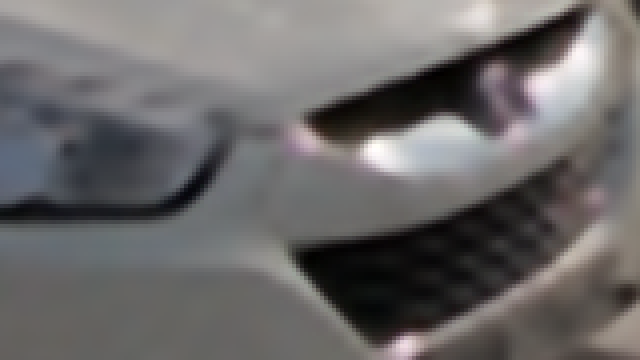}
\footnotesize{ (v1.2) w/o ITA }
\end{minipage}
\begin{minipage}[b]{0.19\textwidth}
\centering
\includegraphics[width=1.0\textwidth]{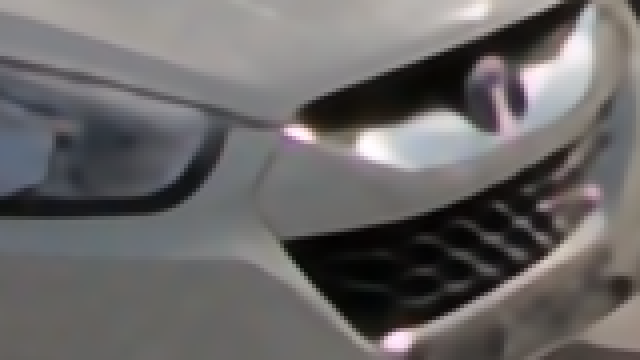}
\footnotesize{ (v1.3) w/o Rec  }
\end{minipage}

\begin{minipage}[b]{0.19\textwidth}
\centering
\includegraphics[width=1.0\textwidth]{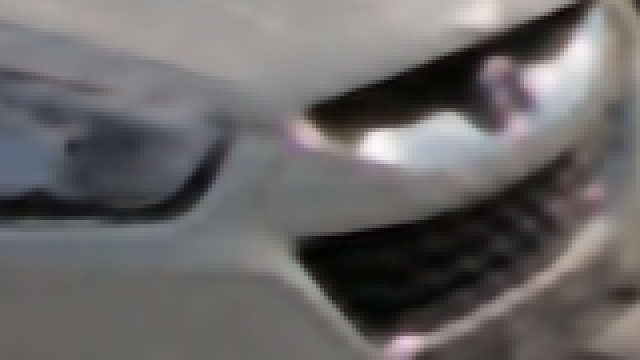}
\footnotesize{ (v1.4) w/o Bi-dir }
\end{minipage}
\begin{minipage}[b]{0.19\textwidth}
\centering
\includegraphics[width=1.0\textwidth]{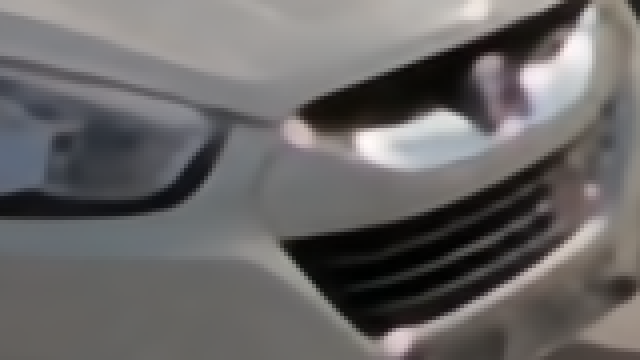}
\footnotesize{ (v1.5) w/o $\mathcal{L}_{C}$  }
\end{minipage}
\begin{minipage}[b]{0.19\textwidth}
\centering
\includegraphics[width=1.0\textwidth]{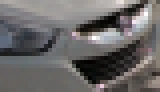}
\footnotesize{ Corrected Frame (Ours) }
\end{minipage}
\begin{minipage}[b]{0.19\textwidth}
\centering
\includegraphics[width=1.0\textwidth]{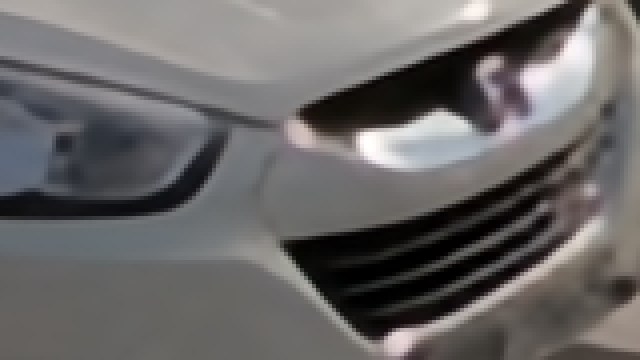}
\footnotesize{ BVSR-IK (Ours) }
\end{minipage}
\begin{minipage}[b]{0.19\textwidth}
\centering
\includegraphics[width=1.0\textwidth]{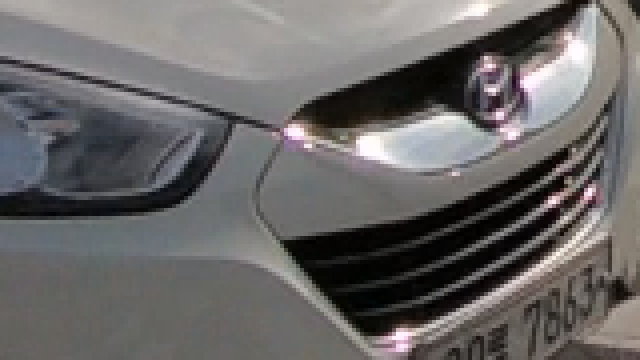}
\footnotesize{ GT  }
\end{minipage}
\caption{{Visualization of ablation studies on BVSR-IK on REDS4~\cite{nah2019ntire} dataset frame for realistic motion blur scenario.}}
\label{fig_Ablation}
\end{figure*}

\noindent {\bf{Effectiveness of the designed modules and loss.}}
We first tested the contribution of our main modules, i.e., ISC module and ITA module,  by creating the following variants.  (v1.1) w/o ISC  - the ISC module was removed from BVSR-IK; (v1.2) w/o ITA - the ITA module was removed from the full model and aligned features were generated by warping with optical flow. We further verified two structures, i.e., recurrent structure in our designed recurrent Transformer and bi-directional propagation in BVSR-IK, resulting in (v1.3) w/o Rec - the hidden state branch was removed from our recurrent Transformer and the corrected frame and warped feature only fed into Transformer; (v1.4) w/o Bi-dir - the ITA module in the backward propagation was removed from our framework; (v1.5) w/o $\mathcal{L}_{C}$ - the correction loss was removed when training our BVSR-IK model. It can be observed from  \autoref{Tab_Module} (a) that our ISC module can bring 0.30 dB PSNR gain and only increase 11.75G FLOPs complexity, and the ITA module achieves 0.62 dB PSNR gain with a 85.75G FLOPs increase. Recurrent structure, bidirectional propagation and correction loss can contribute 0.28 dB, 0.19 dB and 0.08 dB PSNR gains, respectively. We also illustrate the visual results of the above variants and our full model in \autoref{fig_Ablation}.  These results present the visual quality improvement when using our designed modules and loss.

\noindent {\bf{Comparison with existing kernel predictions.}} To compare our designed kernel dictionary with existing kernel prediction methods. We implemented two kernel predictions in BVSR methods, i.e., Self-BVSR~\cite{bai2024self} and FMA-Net~\cite{youk2024fma}, and an unsupervised kernel prediction method in BISR method, i.e., DKP \cite{yang2024dynamic} to replace our designed kernel dictionary in our BVSR-IK model. The corresponding results were reported in \autoref{Tab_Module} (b). Since Self-BVSR \cite{bai2024self} and FMA-Net~\cite{youk2024fma} only predicted one kernel for one frame, the diversity of kernel representation was limited. Additionally, DKP~\cite{yang2024dynamic} improved the kernel representation by a Markov chain Monte Carlo sampling process on random kernel distributions, but did not consider the multi-scale kernel distributions, resulting in insufficient kernel representation and further information exploration.

\noindent {\bf{Comparison with existing correction filters.}} To compare the existing correction filters in our ISC module. We replaced our ISC module with three representative correct filters \cite{hussein2020correction,zhou2023learning,liu2024motion} in our BVSR-IK model and obtained (v3.1) Corr Filter~\cite{hussein2020correction},  (v3.2) Corr Filter~\cite{zhou2023learning} and (v3.1) MISFilter~\cite{liu2024motion}. The results in \autoref{Tab_Module} (c) have confirmed the superior performance of our ISC module over others.

\begin{table}[t]
\setlength{\tabcolsep}{0.500mm}
\fontsize{7}{9}\selectfont
\centering
\resizebox{\linewidth}{!}{\begin{tabular}{c|c|c||c|c|ccccc}
\toprule
$R$ &  {PSNR$\uparrow$ / SSIM$\uparrow$ / tOF$\downarrow$}  & Param.(M) & $N$  &  { PSNR$\uparrow$ / SSIM$\uparrow$ / tOF$\downarrow$}  & Param.(M) \\
\midrule
4  & 29.16 / 0.8105 / 1.95       & 9.10  & 2     & 29.21 / 0.8263 / 1.94   & 6.59  \\
5  & 29.24 / 0.8303 / 1.92       & 9.18  & 4     & 29.30 / 0.8285 / 1.89   & 7.45  \\
6  & 29.37 / 0.8278 / 1.86       & 9.24  & 6     & 29.42 / 0.8291 / 1.86   & 8.39  \\
7  & \bf{29.48 / 0.8303 / 1.82}  & 9.30  & 8     & 29.48 / 0.8303 / 1.82   & 9.30  \\
8  & 29.44 / 0.8307 / 1.81       & 9.37  & 10    & 29.50 / 0.8309 / 1.80   & 10.20  \\
9  & 29.42 / 0.8291 / 1.83       & 9.44  & 12    & \bf{29.53 / 0.8318 / 1.78}   & 11.11  \\
\bottomrule
\end{tabular}}
\caption{{Ablation study of the kernel scale $R$ and atom number $N$.}} \label{Tab_kernel}
\end{table}

\begin{figure}[t]
\centering
\begin{minipage}[b]{0.475\textwidth}
\includegraphics[width=1\textwidth]{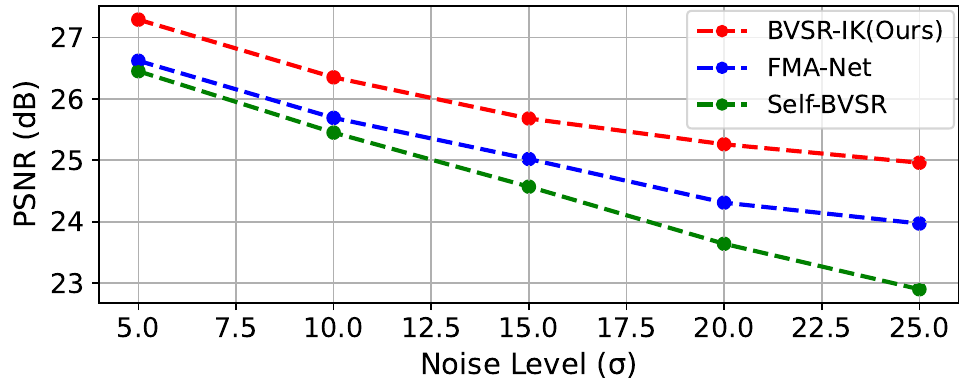}
\end{minipage}
\caption{{Robustness of BVSR models for noise scenario.}}
\label{fig_noise}
\end{figure}


\noindent {\bf{Comparison with existing alignment modules.}} To compare our designed ITA module with the existing alignment module, we used four recent alignments, i.e., flow-based~\cite{chan2021basicvsr}, DCN-based~\cite{tian2020tdan}, FGDA-based~\cite{chan2022basicvsr++}, and FGDF-based \cite{youk2024fma} to replace our ITA module in our BVSR-IK model. Four corresponding variant models were created: (v4.1) Flow, (v4.2), DCN (v4.3) FGDA, and (v4.4) FGDF, respectively. The results in~\autoref{Tab_Module} show that our ITA module has a similar performance to the FGDA module, but with lower complexity.

\noindent {\bf{Effectiveness of kernel scale and atom number.}} To confirm the kernel dictionary scale $R$ and atom number $N$ in our multi-scale kernel dictionary, we set the different $R$ and $N$, respectively, to construct the BVSR model. The corresponding results are presented in \autoref{Tab_kernel}. The max kernel size in kernel dictionary is calculated by $k_{max}$ = $2R$ - 1. We found that with $R$ increasing, the BVSR performance was improved. When larger than $R$ = 8 ($k_{max}$=15), the performance was dropped, which may be due to a large kernel causing the extra blur information. In addition, as $N$ increases, the performance improves, but the model complexity also increases. To avoid excessive model complexity and model size, we set $N$ = 8 in our work.

\begin{table}[t]
\setlength{\tabcolsep}{2.0mm}
\fontsize{7}{9}\selectfont
\centering
\resizebox{\linewidth}{!}{
\begin{tabular}{l|c|c|ccccccc}
\toprule
Noise &  Self-BVSR~\cite{bai2024self} & FMA-Net~\cite{youk2024fma}  & BVSR-IK (Ours) \\
\midrule
$\sigma$=5  &   26.45 / 0.6927 / 3.69    &   26.62 / 0.7052 / 3.62 &   27.29 / 0.7390 / 3.60 \\
$\sigma$=10  & 25.46 / 0.6125 / 4.89    & 25.69 / 0.6567 / 4.46  & 26.35 / 0.7039 / 4.41 \\
$\sigma$=15  &   24.57 / 0.5332 / 4.94    &   25.02 / 0.5851 / 4.90  &   25.68 / 0.6782 / 4.89   \\
$\sigma$=20   & 23.64 / 0.4619 / 5.24   & 24.31 / 0.5300 / 5.21  & 25.26 / 0.6592 / 4.98 \\
$\sigma$=25   &   22.90 / 0.4031 / 5.46   &   23.97 / 0.5252 / 5.28  &   24.96 / 0.6427 / 5.07 \\
\bottomrule
\end{tabular}}
\caption{{Verify the robustness of the BVSR models.}} 
\label{Tab_Noise}
\end{table}

\noindent {\bf{Robustness of the BVSR-IK.}} To confirm the robustness of BVSR models on the noise scenario, we retrain Self-BVSR~\cite{bai2024self}, FMA-Net~\cite{youk2024fma} and our model on the realistic motion blur scenario with noise.  The noise is randomly added in each LR blurred video frame, and the range was set to [0, 25].  We added five noise levels, i.e., 5, 10, 15, 20, and 25, on the REDS4 dataset to evaluate the model robustness. The corresponding results are reported in \autoref{Tab_Noise} and \autoref{fig_noise}. It is found from these results that our BVSR-IK has better robustness for the noise scenario compared with self-BVSR and FMA-Net, when the noise level increases. 


\section{Conclusion}
In this paper, we proposed a novel blind video super-resolution framework, BVSR-IK, which is the first attempt to employ INR to generate a multi-scale kernel dictionary for spatially varying degradations in blind super-resolution. Based on the implicit kernel dictionary, a new recurrent Transformer is designed to predict coefficient weights by capturing both short-term and long-term scale-aware information, which are then employed for accurate filtering for implicit spatial correction and implicit temporal alignment module. Our model was evaluated on commonly used datasets and achieved consistent and evident performance gains over SOTA BVSR methods.

\section*{Acknowledgments}
This work was supported by the Natural Science Foundation of Sichuan
Province under Grant 2023NSFSC1972, the China Scholarship Council, the University of Bristol, and the UKRI MyWorld Strength in Places Programme (SIPF00006/1).



{
\small
\bibliographystyle{ieeenat_fullname}
\bibliography{main}
}

\end{document}